\documentclass[letterpaper, 10 pt, conference]{ieeeconf}

\IEEEoverridecommandlockouts                              

\usepackage{xcolor}

\definecolor{MyDarkBlue}{rgb}{0,0.08,1}
\definecolor{MyDarkGreen}{rgb}{0.02,0.6,0.02}
\definecolor{MyDarkRed}{rgb}{0.8,0.02,0.02}
\definecolor{MyDarkOrange}{rgb}{0.40,0.2,0.02}
\definecolor{MyPurple}{rgb}{111,0,255}
\definecolor{MyRed}{rgb}{1.0,0.0,0.0}
\definecolor{MyGold}{rgb}{0.75,0.6,0.12}
\definecolor{MyDarkgray}{rgb}{0.66, 0.66, 0.66}

\DeclareRobustCommand{\slnote}[1]{\textcolor{MyRed}{\textbf{[Sandra: #1]}}}

\usepackage{graphics} 
\usepackage{epsfig} 
\usepackage{mathptmx} 
\usepackage{times} 
\usepackage{amsmath} 
\usepackage{amssymb}  
\usepackage{color}
\usepackage{gensymb}
\usepackage{siunitx}

\newcommand{\myparagraph}[1]{\vspace{0.1in}\noindent\textbf{#1}}

\title{\LARGE \bf
Exoskeleton-covered soft finger with vision-based proprioception and tactile sensing
}

\author{
    \authorblockN{Yu She$^{1*}$, Sandra Q. Liu$^{1*}$, Peiyu Yu$^{1,2*}$ and Edward Adelson$^{1}$}
        \authorblockA{$^{1}$Massachusetts Institute of Technology, $^{2}$Tsinghua University\\
    {\tt\small <yushe,sqliu,yupy>@mit.edu, adelson@csail.mit.edu}} 
\thanks{* Authors with equal contribution.}%
}

\usepackage{multirow}
\usepackage{comment}

\begin{document}

\maketitle
\thispagestyle{empty}
\pagestyle{empty}

\begin{abstract}
Soft robots offer significant advantages in adaptability, safety, and dexterity compared to conventional rigid-body robots. However, it is challenging to equip soft robots with accurate proprioception and tactile sensing due to their high flexibility and elasticity. In this work, we describe the development of a vision-based proprioceptive and tactile sensor for soft robots called GelFlex, which is inspired by previous GelSight sensing techniques. More specifically, we develop a novel exoskeleton-covered soft finger with embedded cameras and deep learning methods that enable high-resolution proprioceptive sensing and rich tactile sensing. To do so, we design features along the axial direction of the finger, which enable high-resolution proprioceptive sensing, and incorporate a reflective ink coating on the surface of the finger to enable rich tactile sensing. We design a highly underactuated exoskeleton with a tendon-driven mechanism to actuate the finger. Finally, we assemble 2 of the fingers together to form a robotic gripper and successfully perform a bar stock classification task, which requires both shape and tactile information. We train neural networks for proprioception and shape (box versus cylinder) classification using data from the embedded sensors. The proprioception CNN had over 99\% accuracy on our testing set (all six joint angles were within 1$^\circ$ of error) and had an average accumulative distance error of 0.77 mm during live testing, which is better than human finger proprioception.  
These proposed techniques offer soft robots the high-level ability to simultaneously perceive their proprioceptive state and peripheral environment, providing potential solutions for soft robots to solve everyday manipulation tasks. We believe the methods developed in this work can be widely applied to different designs and applications.

\end{abstract}

\section{INTRODUCTION}


Compared to traditional rigid robots, soft robots offer significant advantages in safety, adaptability, dexterity, and robustness \cite{trivedi2008soft}, and have demonstrated unique capabilities in the manipulation of gentle, fragile, and various objects with different shapes and sizes \cite{shintake2018soft, truby2019soft}. Due to their elastic bodies, soft robots can passively adapt to objects they interact with, allowing them to have safe interactions with their environment and to have intrinsic robustness to uncertainty. As a result, soft robots have widespread potential applications in surgical robots and in prosthetics, in safe human-robot interactions, and in numerous other applications \cite{she2016modeling,gorissen2017elastic,wang2018toward}. 

\begin{figure}[h]
	\centering
	\includegraphics[width=1.0 \linewidth]{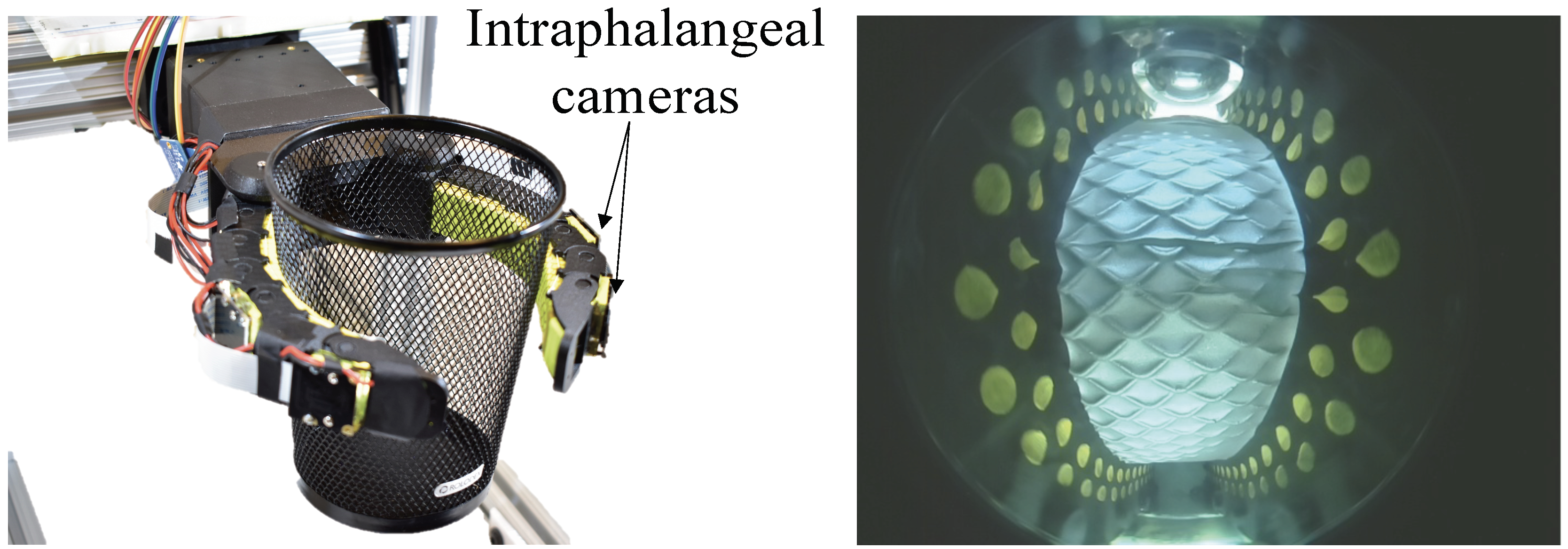}
    \vspace{-20pt}
	\caption{\textbf{Left} The robotic gripper is holding a mesh cup. The gripper is comprised of 2 exoskeleton-covered soft fingers, and each finger is embedded with 2 intraphalangeal cameras. \textbf{Right} The corresponding image from an embedded camera. The dot features on either side are designed for soft robot proprioception and the painted layer in the middle is designed for soft robot tactile sensing.  
}
	\label{fig:introductionofthesoftfinger}
\end{figure}

Soft robots should be equipped with rich sensory perception abilities akin to those of living creatures, especially when they are employed to prosthetic applications \cite{zhao2016optoelectronically}. Furthermore, embedding sensors in soft robots could significantly improve manipulation performance \cite{shintake2018soft} and enhance the robustness of the robot interaction with environment \cite{van2018soft}. Particularly, soft robots need to be capable of perceiving their own state (proprioception) in addition to external stimuli (tactile sensing) to perform tasks as smoothly as biological systems \cite{wang2018toward}. However, due to the high flexibility of soft robots, soft robot proprioception and tactile sensing remain considerable challenges that significantly limit their applications \cite{truby2019soft,wang2018toward,wang2019real}. 

Typical solutions for soft robot perception include embedding sensors along neutral axes to measure global shape of soft robots \cite{van2018soft, case2015soft, she2015design,zhao2016helping} and mounting sensors on robot surfaces to estimate pressure and touch \cite{park2007force,wang2009thin}. To some degree, these types of sensors provide feasible ways for soft robot perception. However, they can only afford limited sensing data, which cannot fully capture the infinite degree of freedom soft robots can achieve. We still need to significantly increase the dimensionality of sensing data in order to fully understand the shape and cutaneous states of soft robots.

To solve this problem, we introduce a novel proprioceptive and tactile sensor, GelFlex, which leverages embedded cameras to simultaneously obtain proprioceptive and tactile data for soft robots. Embedded cameras are able to provide rich and powerful high-resolution information which is highly synergistic with recent image-based deep learning techniques \cite{donlon2018gelslim}. Wang~\textit{et al}.~\cite{wang2019real} explored the usage of embedded cameras for sensing the shape of soft robots. However, their group focused on perception and the soft robots that they studied cannot perform grasping or manipulation tasks because the camera is installed at the base of the robot. This camera position only allows a small sensing range of robot deformation, which is not fit for soft robot proprioception because large deformations are essential and inevitable in soft robot grasping and manipulation tasks.

Here, we develop a robotic gripper comprised of 2 exoskeleton-covered soft fingers, each embedded with 2 cameras, as shown in Fig.~\ref{fig:introductionofthesoftfinger}. We design features on both sides of the finger and we assemble the cameras into the finger body such that the cameras have a wide visibility range even with large deformations of the finger. Regular testing shows that a single camera can sense 105$^{\circ}$ of bending in our soft finger. In addition, we design a semi-specular paint layer at the bottom surface of the finger and illuminate the finger body using LEDs assembled on the top surface of the finger. To the authors' knowledge, this novel design allows us to obtain vision-based propioception and tactile sensing in soft robots for the very first time. We also design a highly underactuated exoskeleton with a cable-driven system to actuate the gripper to perform tasks. Finally, we train neural networks for proprioception and objects classification. Specifically, we tested the algorithms on differentiating various shapes and sizes of bar stock, a classification problem which simultaneously needs shape and tactile information. 

Altogether, our work present the following contributions:
\begin{itemize}
    \item A novel design of a highly under-actuated exoskeleton-covered soft finger with a tendon-driven system;
    \item A first design of a fully-functioning gripper with embedded vision sensors for proprioception, tactile sensing, and both simultaneously;
    \item A neural network for estimating the shape of the finger with the sensory data from the embedded cameras;
    \item A neural network that can classify size and profile of round/square bar stock based on the data obtained from the embedded cameras.
\end{itemize}

\section{RELATED WORK}
\subsection{Soft robot proprioception} \label{review1}
Embedded sensors along neutral axes are widely used for measuring specific types of deformation in soft robots. For instance, Homberg~\textit{et al}.~\cite{homberg2019robust} explored robust proprioceptive grasping of a soft robot hand with embedded resistive force sensors on fingertips and resistive bend sensors along the finger. Soter~\textit{et al}.~\cite{soter2018bodily} integrated 4 bending sensors in a soft octopus-inspired arm for propriocetive sensing.
However, these types of sensors may suffer from hysteresis, which can potentially be overcome by optical sensors. Zhao~\textit{et al}.~\cite{zhao2016optoelectronically} investigated stretchable optical sensors for strain sensing for a soft robotic hand. Van Meerbeek~\textit{et al}.~\cite{van2018soft} developed soft proprioceptive foam using optical fiber and machine learning techniques. Molnar~\textit{et al}.~\cite{molnar2018optical} studied the tip orientation and location in 3D space of pneumatically-controlled soft robots using optical sensors and neural networks. 

The latest exploration of proprioception for soft robots covers novel material and sensing techniques. Truby~\textit{et al}.~\cite{truby2019soft} integrated ionogel soft sensors with 3D printing techniques to develop soft robotic fingers with proprioceptive and tactile sensing. Thuruthel~\textit{et al}.~\cite{thuruthel2019soft} investigated soft robot perception with embedded cPDMS sensors and general machine learning approaches. But, these aforementioned sensors only provide low dimensional sensory data for soft robot proprioception. On the other hand, vision-based sensors can generate powerful and rich sensing information which potentially correspond well to the high dimensionality of deformation in soft robots. Wang~\textit{et al}.~\cite{wang2019real} started to introduce embedded vision sensors for soft robot proprioception. They proposed a framework to predict real-time 3D shapes of robots using a neural network. However, the way they designed the robot significantly limits the sensing range of the robot and hence substantially restricts its application.

\subsection{Soft robot tactile sensing} \label{review2}
Soft electronic skins have attained remarkable progress in the past few decades. Most of them are flexible and stretchable, making them potentially viable for soft robot tactile sensing \cite{wang2018toward}. The integrated ionogel soft sensors in \cite{truby2019soft} can estimate tactile information and proprioception for soft robots. However, these methods may suffer from low dimensional sensory data for high-resolution tactile sensing. Again, vision-based sensors may provide a simple but effective solution for tactile sensing. Soter~\textit{et al}.~\cite{soter2018bodily} reported work including both proprioception and exteroception of a soft robot, where they used visual data from a global camera for exteroceptive representation of their soft robot. However, using the global camera might make it difficult to sense tactile data of the soft robots. We believe that embedded cameras may be one of the possible solutions to address high resolution tactile sensing for soft robots. Currently, there are a variety of vision-based sensors for tactile sensing for rigid robotic hands, but there are none for soft robot tactile sensing, to the best knowledge of the authors. 

One type of vision-based tactile sensor is the GelSight sensor \cite{dong2017improved}. In their work, they applied a soft gel with painted patterns on the gel surface as a sensing medium and then used embedded cameras to track the high-resolution tactile information. The technique for tactile sensing in this work is similar to the technique of the GelSight sensor. However, where GelSight sensors have predominantly been used for rigid body robot applications, we used it for a soft robot system.

\section{METHOD}

\subsection{Hardware}

\begin{figure}[t]
	\centering
	\includegraphics[width=1.0\linewidth]{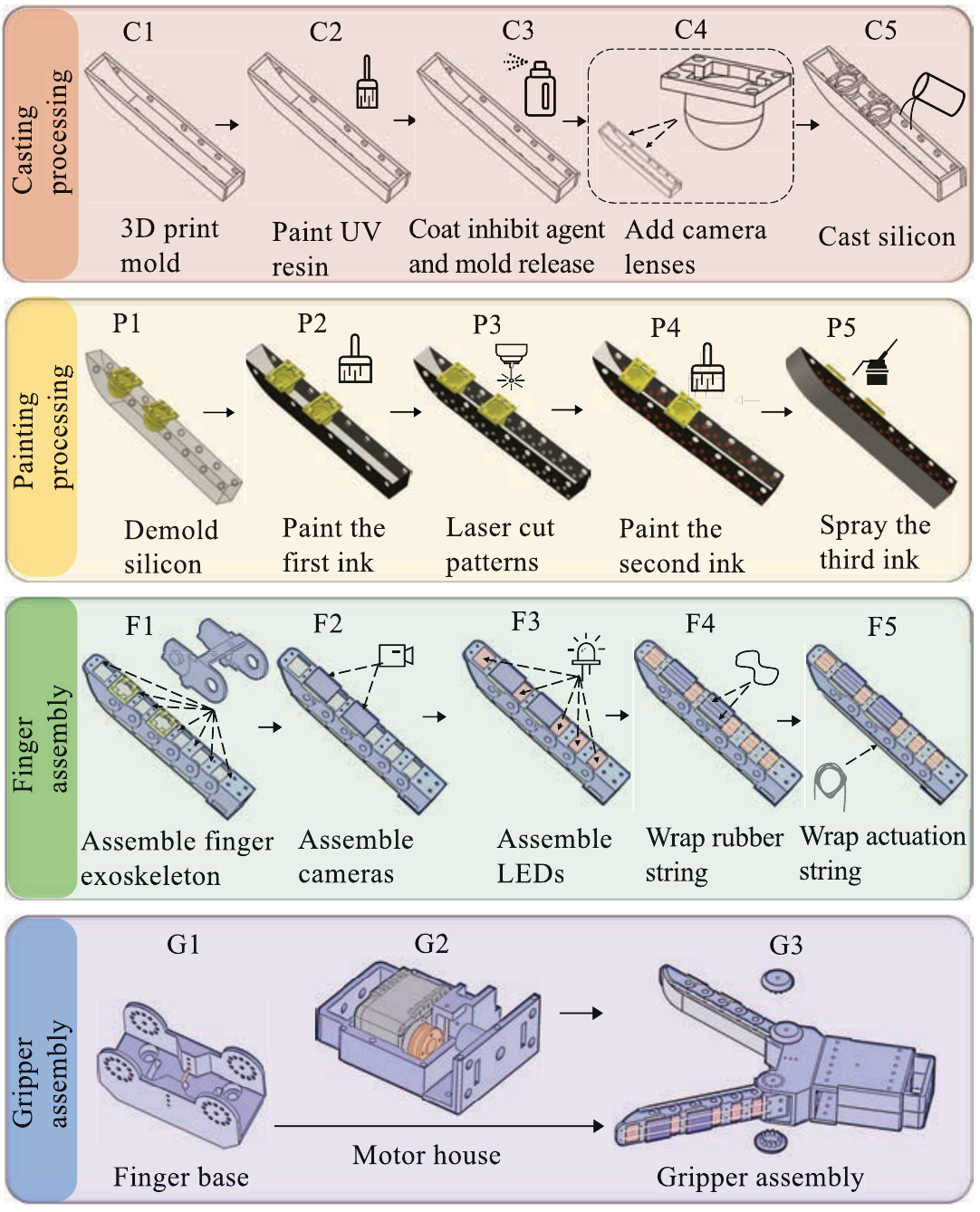}
    \vspace{-15pt}
	\caption{Process flow for manufacturing of a robotic gripper including casting processing, painting processing, finger assembly, and gripper assembly.}
	\label{fig:Process flow for manufacture}
	 \vspace{-15pt}
\end{figure}

The manufacturing of the robotic gripper includes casting processing, painting processing, finger assembly, and gripper assembly, and the overall process flow for manufacturing of the soft gripper is shown in Fig.~\ref{fig:Process flow for manufacture}.

\myparagraph{Casting processing} We first make the finger mold  using a 3D printer (Onyx One 3D Printer, Markforged, Inc.) $\{$C1$\}$. The finger surfaces must be smooth since both the tactile and shape information are acquired via cameras observing these surfaces from inside. To improve the smoothness, we paint the inside of the 3D-printed finger mold  with ultraviolet (UV) resin (200g Crystal Clear Ultraviolet Curing Epoxy Resin, LIMINO) $\{$C2$\}$. After exposing the resin to UV lights for 1 hour, the top of the mold becomes smooth. However, the UV resin contains chemicals which prevent the curing of silicone. To counter this, we coat the UV resin with a layer of an inhibit agent (Inhibit X, Smooth-On, Inc.), and then spray a layer of mold release  (Ease Release $^{\text{TM}}$ 200, Smooth-On, Inc.) over it $\{$C3$\}$. We then embed clear acrylic hemisphere lenses (3/4 inch Hemispheres, California Quality Plastics) in the finger to allow for the integration of fish eye camera lenses $\{$C4$\}$. These hemispheres provide wide view angles for the camera and hide the camera within the finger. Since we use cameras for both proprioception and tactile sensing, the soft finger body must be optically transparent. To give our finger this property, we made it out of platinum-catalyzed translucent silicone (XP-565, Silicones, Inc.) $\{$C5$\}$. Prior to adding the silicone, we apply primer (DOWSIL $^{\text{TM}}$ PR-1200 RTV Prime Coat, DOW) on the outside surface of the lens to bond it to the silicone. This procedure prevents the cured silicone from easily being ripped off during larger deformations of the finger. To improve the softness and resilience of the silicone finger, we add dilute agent (LC1550 Phenyl Trimethicone, Lotioncrafter). The ratio of the components of silicon XP-565 A and B, and phenyl trimethicone is 1 to 15 to 3. After uniformly mixing all the components in a plastic cup, we vacuum it (90062-A 3 CFM Vacuum Pump, Mastercool, Inc.) to get rid of bubbles. Finally, we pour the mixture into the 3D printed mold. 



\myparagraph{Painting processing}
We first de-mold the silicone from the 3D printed mold and clean any residue on the silicone using soap $\{$P1$\}$. We then use different colors of silicone inks (Print-On Silicone Ink, Raw Material Suppliers), which are scratch-resistant and durable, to create features. Silicone solvent (NOVOCS Gloss, Reynolds Advanced Materials) is added to the silicone ink to dilute the ink. The ratio we use is 1 to 10 to 30 (silicone ink catalyst to silicone ink to silicone solvent). We apply black ink on both sides of the silicone finger and cure it in an oven at 200$^\circ$F for 30 minutes $\{$P2$\}$. Then, we use a laser cutter (PLS6.75, Universal Laser Systems, Inc.) to engrave circular patterns on both sides of the finger and remove the black paint in those areas $\{$P3$\}$. Next, we paint yellow silicone ink on both sides of the finger, creating yellow dots with a black background $\{$P4$\}$. These yellow dots allow the camera to more easily capture finger deformation information. The next step is to spray a reflective ink at the bottom surface of the finger for tactile sensing $\{$P5$\}$. We opted to use semi-specular painting material (silver silicone ink) so that we could observe a higher-level of intricate details. Afterwards, we use an ultrasonic processor (VCX 750, Sonics $\&$ Materials, Inc.) to enhance the dissolution of the silicone ink into the NOVACS solvent. We use a airbrush (ZENY Pro 1/5 HP Airbrush Air Compressor Kit, ZENY) to paint the silver silicone ink on the bottom surface of the finger to ensure a smooth and uniform paint finish. Then we place the painted finger in an oven for 2 hours at 200$^\circ$F before the finger is ready to use. Note that the painting order of the silicone ink can vary as long as one can create the desired features. 


\myparagraph{Finger assembly}
We design a highly underactuated exoskeleton that is comprised of 7 segments and 6 joints to actuate the soft finger. The exoskeleton is then integrated with the soft silicone trunk. We first make the finger exoskeleton with the 3D printer and assemble the finger segments, before pressing the soft silicone trunk into the exoskeleton $\{$F1$\}$. This design allows the finger to easily adapt to grasping various objects. The exoskeleton is designed with round pegs on the side to fit into the round slots on the side of the soft trunk. This prevents the soft trunk from easily detaching from the exoskeleton. Then, we add two cameras to the camera lenses that were fixed onto the silicone finger during the casting processing $\{$F2$\}$. Next, we add white LEDs, which provide internal illumination for the camera $\{$F3$\}$, and restrain them with rubber bands attached to the top of the exoskeleton $\{$F4$\}$. Finally, we add actuation strings through holes on each segment of the exoskeleton $\{$F5$\}$. One end of the string is tied to the last exoskeleton segment and the other end of the string is attached to a motor disk. This yields a highly underactuated exoskeleton-covered soft finger with a tendon-driven system. 
Note, here the soft silicon trunk with its surface painting as well as the associated cameras and LEDs are referred to the \textit{GelFlex}.


\myparagraph{Gripper assembly}
The last step of the hardware process is to assemble the gripper. The gripper includes assembling 2 GelFlex fingers, a finger base $\{$G1$\}$, and a housing $\{$G2$\}$ for a servo motor (DYNAMIXEL XM430-W210-T, ROBOTIS) and disk. The embedded cameras are controlled using Raspberry Pi boards. The orientation of the finger relative to the finger base is adjustable $\{$G3$\}$, allowing us to use it for a variety of tasks.

\subsection{Software}
We use a combination of three neural nets as a solution to this problem. One was for proprioception of the GelFlex finger while the other was to determine, through tactile sensing, whether the object being grasped was a box or cylinder. The last neural net combined the shape information of the finger and the tactile knowledge to classify the size of the shape that was being grabbed.

\myparagraph{Proprioception}
To obtain proprioception of the finger, we utilized the 2-dimensional position of the individual joints of the finger exoskeleton. This allowed us to simplify the learning problem and use the mechanical constraints of the exoskeleton to reduce the dimensionality of the soft robot. 

\begin{figure}[h]
	\centering
	\includegraphics[width=0.7\linewidth]{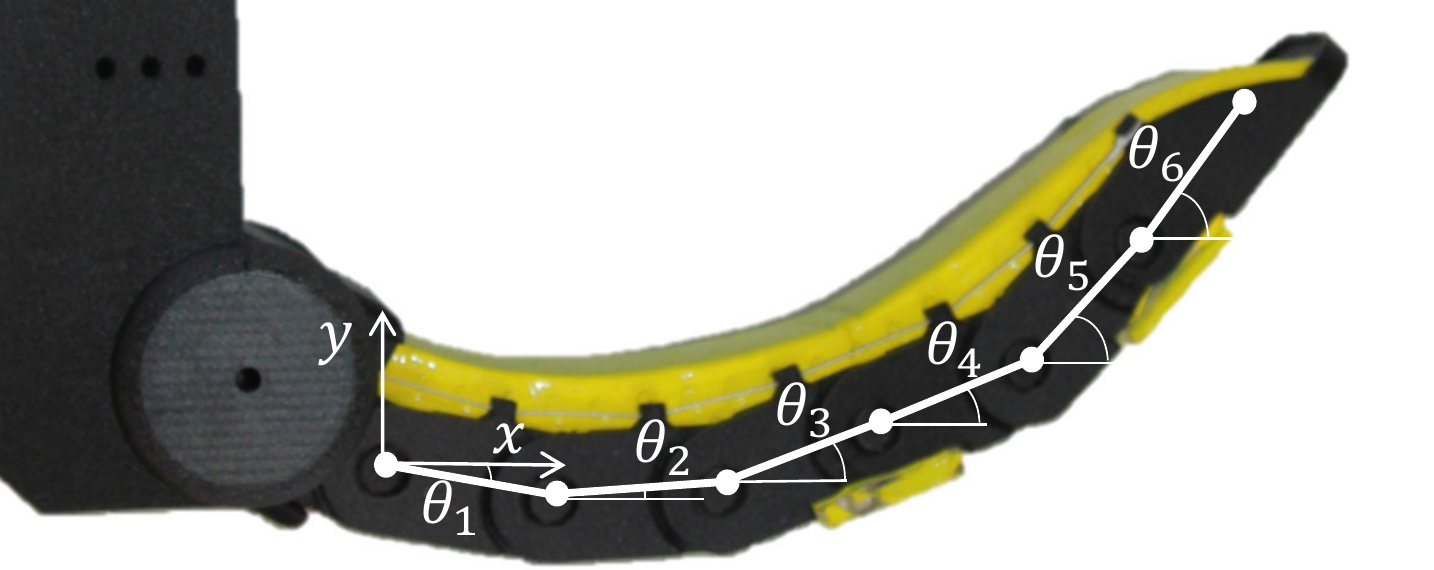}
    \vspace{-5pt}
	\caption{Convention for calculating and determining angles from the finger exoskeleton.}
	\label{fig:abs_ang}
\end{figure}

Taking the first stationary joint as the world reference frame origin, we calculated the absolute angle (relative to the world reference frame) of each joint for a total of six numbers (Fig. \ref{fig:abs_ang}). 

We collected approximately 3500 images from each of the cameras inside of the finger and the corresponding finger position absolute angles. These images include ones where the gripper is empty and ones where the gripper is grabbing arbitrary objects. We used this method to collect a wider variety of possible finger positions and to allow the neural net we trained to parse images with tactile information. We used 80\% of these images for training and retained 20\% for testing. We trained one neural network using the camera in the middle of the finger as a singular input and we trained another neural network using both of the embedded camera images. 

\begin{figure}[h]
	\centering
	\includegraphics[width=1\linewidth]{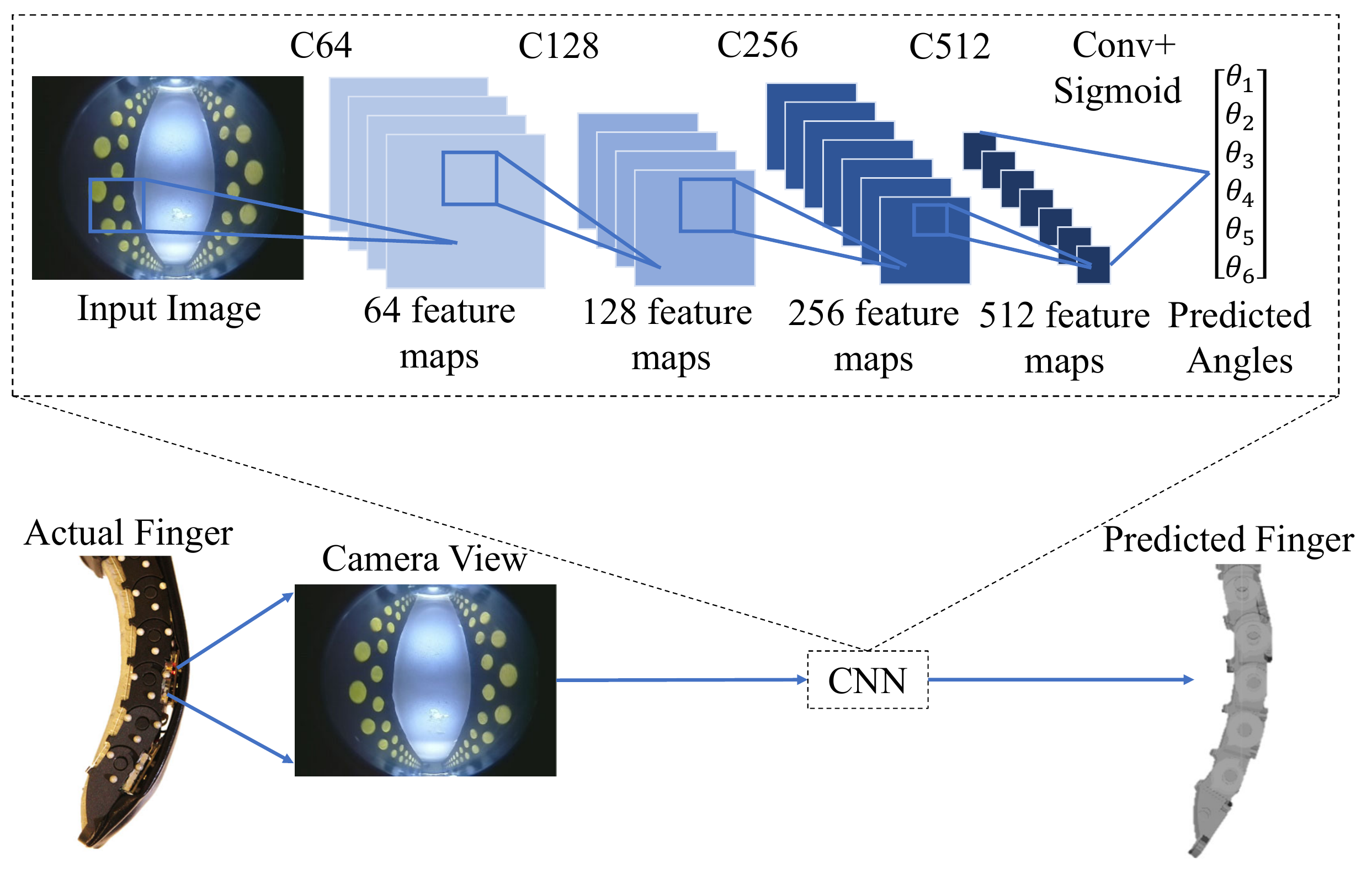}
	\caption{\textbf{Top} Architecture of the fully convolutional network. $Ck$ denotes a Convolution-BatchNorm-ReLU layer with k filters. \textbf{Bottom} Global view of finger paired with the embedded camera view, which gets fed into the CNN. The CNN outputs values that correspond to a computer simulation of the finger (we use RViz, which is a robot visualization package in Robot Operating System/ROS \cite{ROS}).}
	\vspace{-10pt}
	\label{fig:deg_net}
\end{figure}

\textit{ a) Network Architecture:} Given that the inputs are 2D images, we leverage the fully convolutional neural network (CNN)~\cite{Long2015fcn} to process the data. The architecture of the network we use for single image is a sequential model of 4 convolutional layers with batch normalization and ReLU, followed by a single convolutional layer with sigmoid activation that maps the extracted feature to the angle vector (see Fig~\ref{fig:deg_net}). We use a similar architecture for the neural network trained with two images. The only difference is that we concatenate the images from both cameras, and the input channel number is doubled.

\textit{ b) Data Pre-processing and Data Augmentation:} 
Because the ground truth angles have significantly different variances, we normalize the angles using the mean and standard deviation of the training set. During the testing procedure, we re-scale the predicted angles.

We add Gaussian noise with zero mean to the ground truth angles to simulate the possible real-world perturbations. We set the noise variance to 1e-3 to ensure that the ground truth data will not be overwhelmed by the noise. Then, to improve the robustness of the model to illumination changes, we augment the image data by randomly changing the contrast and saturation of the images. 

\begin{figure}[h]
	\centering
	\includegraphics[width=0.8\linewidth]{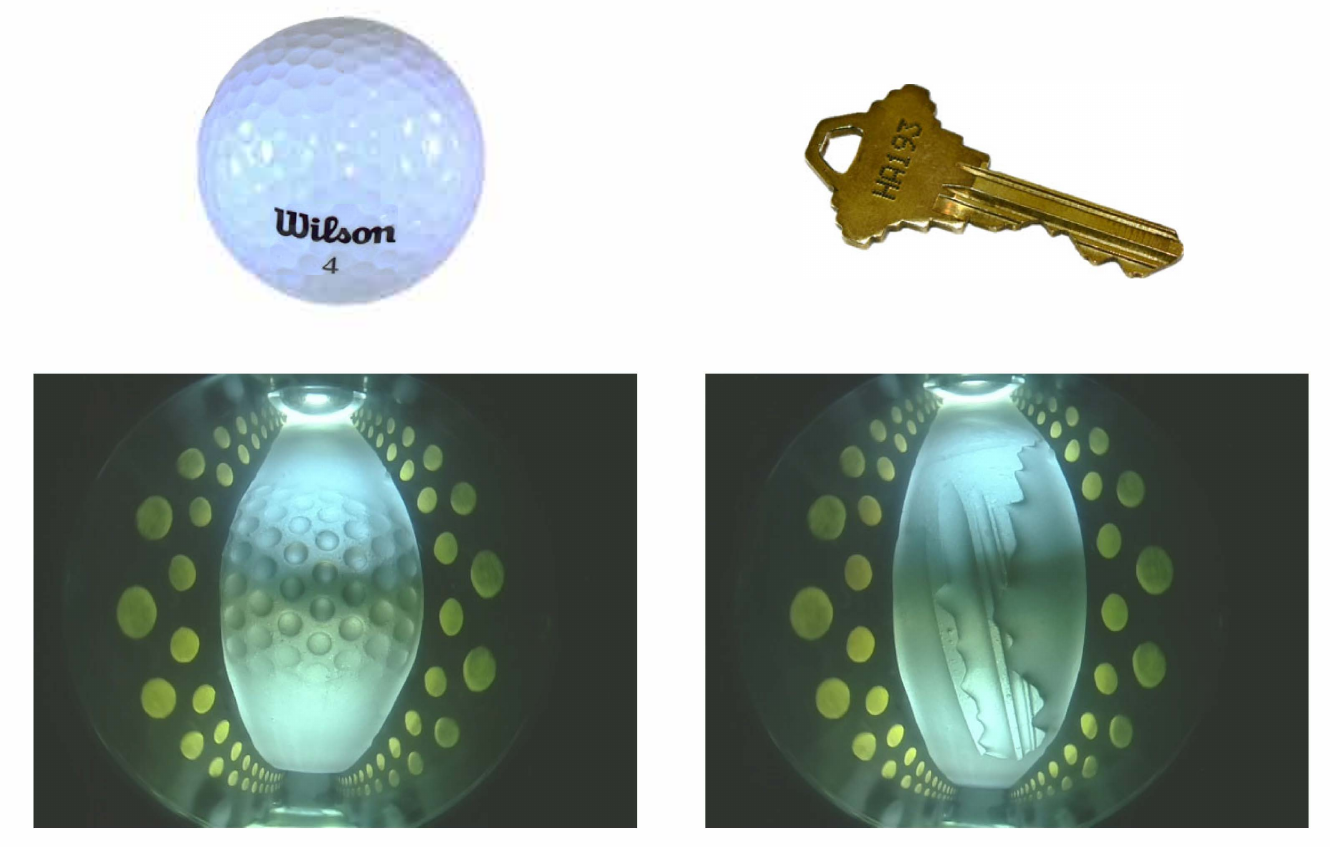}
    \vspace{-10pt}
	\caption{\textbf{First row} Objects that we placed in contact with the finger (from left to right: a golf ball, a key). \textbf{Second row} Tactile imprint of the object viewed via the internal camera.}
	\label{fig:tactile_sense}
\end{figure}

\myparagraph{Tactile Sensing}
Using the center camera embedded in the GelFlex finger, we were able to obtain highly detailed images of objects the silicone finger came into contact with, such as golf balls and keys (Fig. \ref{fig:tactile_sense}). We could also quantitatively observe the difference between the finger touching sharp corners (i.e. boxes) versus it touching smoother surfaces (i.e. cylinders).

To place emphasis on the tactile imprint in our images, we took the difference images of the tactile contact images and a calibration image. The calibration image was one of the finger without tactile imprints at its non-actuated starting position. Then, to minimize the effect the different bending patterns the finger images have, we crop the tactile information.

We collected over 850 images of the finger contacting various sizes of box and cylindrical objects (half of each). The contact points were varied, as were the orientations of the objects. These images were used as inputs for training a simple CNN. We retained 25\% of the images for testing and used the remaining 75\% for training.

\textit{ a) Network Architecture:}
The network architecture, which is based on the LeNet-4 model \cite{LeNet}, is composed of a sequential model with two convolutional layers (kernel size 5x5) that include ReLU and max pooling (kernel size 2x2), followed by two fully connected layers. We use ReLU and softmax activation for the last two layers, respectively. 

\myparagraph{Size Estimation}
A total of 800 predicted angles and labels (200 for each class size: 4.25, 4.5, 4.75, 5.0 inches; 400 for each shape: boxes and cylinders) were collected for the neural network. These were the shape sizes that the soft gripper could comfortably and successfully grab. We randomly sampled 90\% of the pairs of data for the training set, and used the rest as the testing set. 

\textit{ a) Network Architecture:} We consider 3 architectures to incorporate the class information (i.e., class label) into the size estimation process: multiple layer perceptron (MLP) model, Two-path MLP model and our proposed approach, \emph{MLP with Neural Incorporator}.

We use the MLP model as our baseline to assess the performance of other architectures because it can generalize well on unseen data. The inputs of the model are the concatenation of predicted angles and class label. 

Although the MLP model can solve the size estimation problem (88\% accuracy on the testing set), we find that the performance can be further improved if we use two-path MLP model (93\% accuracy on the testing set). We use two paths of MLP to separately process the predicted class label and angles, and fuse them in the embedding space by channel-wise concatenation. The combined feature is then fed into a final MLP to perform size estimation.

\begin{figure}[!htbp]
	\centering
	\includegraphics[width=1.0\linewidth]{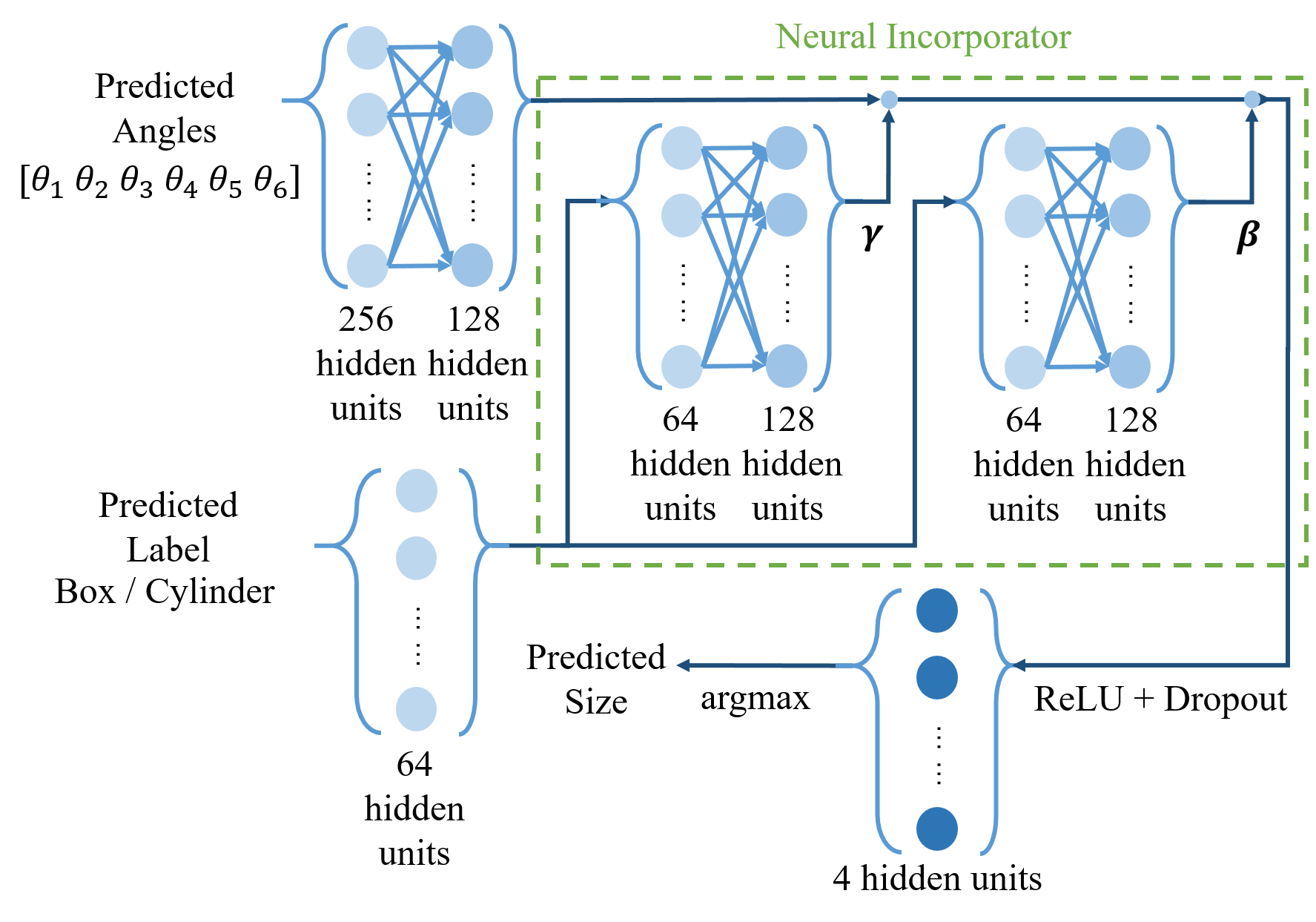}
    \vspace{-20pt}
	\caption{Architecture of the MLP with Neural Incorporator model.}
	\label{fig:size_net_3}
\end{figure}

Next, we propose the Neural Incorporator model to fuse the feature in an element-wise manner and achieve a much better performance (100\% accuracy on the testing set). Inspired by~\cite{Park2019spade}, we use the feature learned from the class label $\mathbf{\beta}, \mathbf{\gamma}$ to normalize the features learned from the predicted angles $\mathbf{f}$ (Fig~\ref{fig:size_net_3}) as follows,
\begin{equation}
    \mathbf{\hat{f}} = \mathbf{f}*(\mathbf{1} + \mathbf{\gamma}) + \mathbf{\beta}
\end{equation}{}

\vspace{-15pt}
\section{EXPERIMENT}

\begin{figure}[h]
	\centering
	\includegraphics[width=1\linewidth]{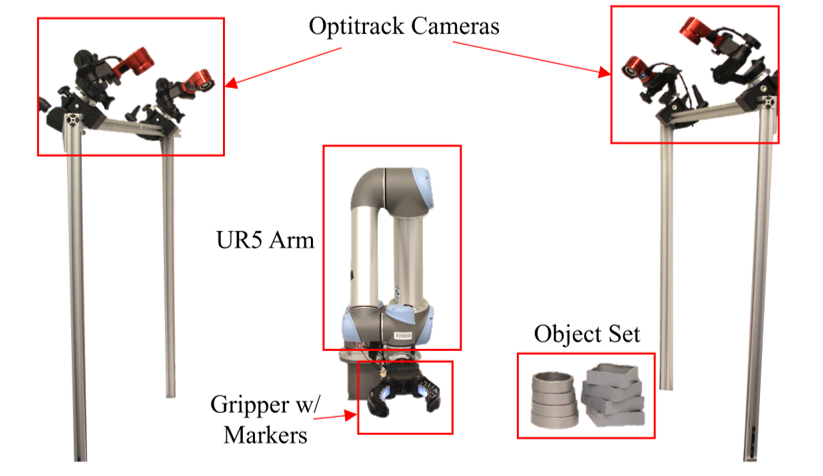}
    \vspace{-10pt}
	\caption{Experimental setup. We attach the soft robot gripper to the end of the UR5 arm to keep it in place. The four OptiTrack cameras keep track of the position of the markers on the finger and we test our gripper using the eight objects in the object set.}
	\label{fig:introduction}
\end{figure}

\subsection{Experimental Setup}
Three 3 mm OptiTrack (OptiTrack, NaturalPoint, Inc.) face markers were adhered onto the surface of each joint in the finger and on the tip of the finger to define the individual positions of each rotation joint. These positions were streamed via the vrpn\_client\_ros while the cameras streamed images via mjpg\_streamer.

Four cylinders and four rectangular prisms (square faces), were created using laser cut acrylic. We used these objects because they are the two most common bar stock profiles that are utilized. The cylinders were of diameters and square face widths of 4.25, 4.5, 4.75, and 5.0 inches, which were the smallest sizes our gripper could grab. We chose to have 0.25 inch (approximately 6.35 mm) differences to demonstrate the ability of our proprioception neural network to differentiate between typical differences in bar stock sizing. We refer to these objects as our object set.

To grab the objects with our gripper fingers, we keep track of the position of the motor. Once it has grasped an object (position reading stays constant), we use our neural nets to determine the size and type of object our gripper is grabbing. Because of gripper symmetry, we choose to only obtain readings from one of its fingers. 

We also note that the classification of shape profile and size of bar stock is not a task that can be accomplished with only proprioception or tactile sensing. Although our current tactile sensing can sense features (i.e. edges, rounded surfaces), we cannot differentiate between the different curvature of the cylinders or similar edge placements of different squares. On the other hand, the finger structure while gripping certain cylinders is similar to its structure while gripping certain box shapes, making it difficult to only use proprioception.

\subsection{Results}
\myparagraph{Proprioception}
For both of neural nets we trained (using one image input versus using two image inputs), we had extremely high accuracy (over 99\% accuracy where all six angles were within 1$^{\circ}$) on our testing set. We then directly compared the neural net predicted angles to the angles calculated using the OptiTrack positional data for live testing. Using the single input neural net, the average sum of the absolute errors in the six angles was $1.29^{\circ}$ while not grasping anything and $2.22^{\circ}$ while grasping objects in the object set. For the double input neural net, these errors were $1.16^{\circ}$ and $4.10^{\circ}$, respectively. 

\begin{figure}[h]
	\centering
	\includegraphics[width=0.8\linewidth]{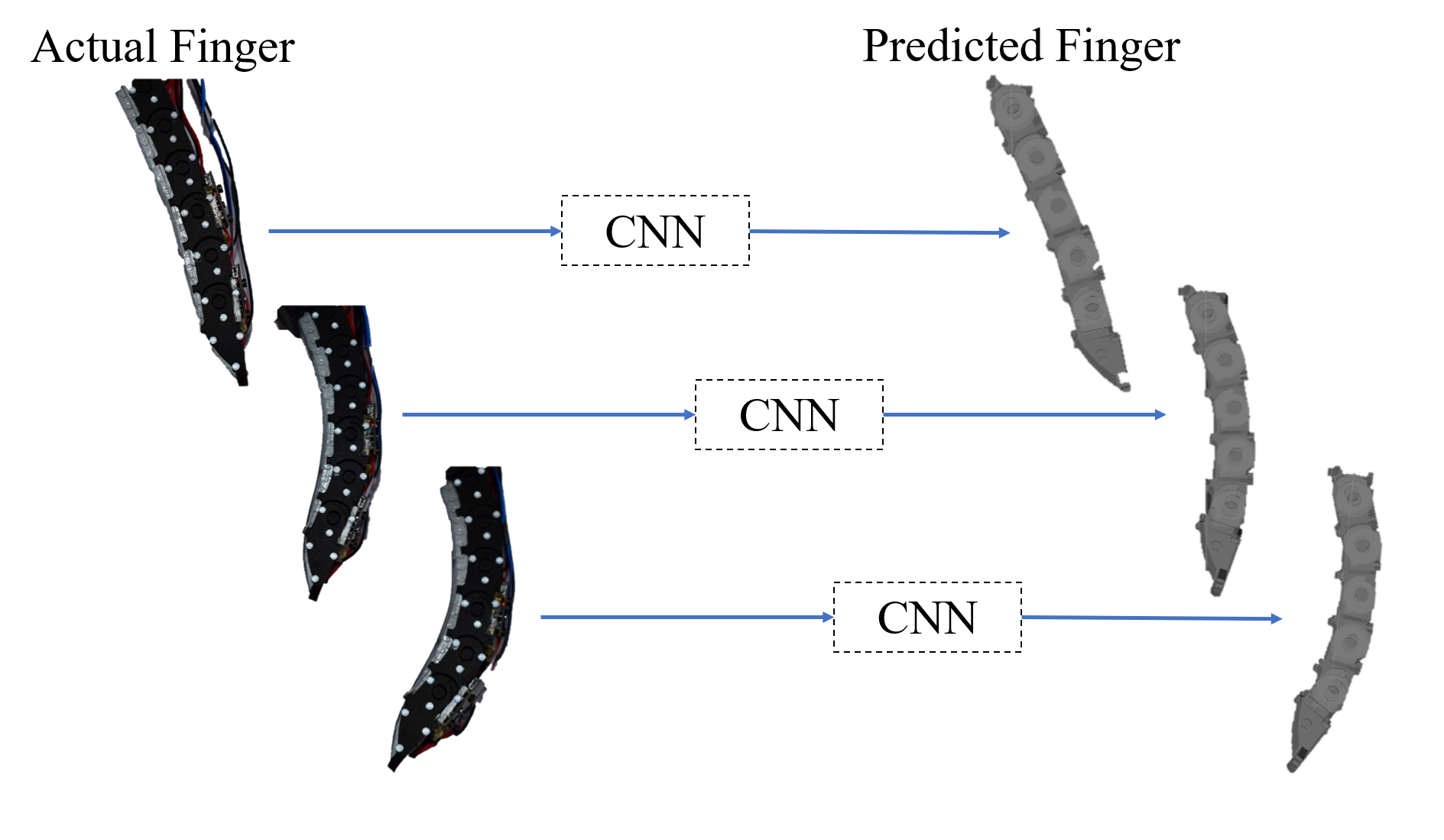}
    \vspace{-20pt}
	\caption{The actual finger positions versus their corresponding predicted fingers (shown in Rviz). We fed the intraphalangeal camera image into the neural net (see Fig. \ref{fig:deg_net} for a description of the CNN) and outputted the angles into Rviz.}
	\label{fig:actual_v_predicted}
\end{figure}

Our position error calculations involved using the ground-truth angles and neural network estimation angles to back-calculate the ground-truth and neural network position value of the last joint in the gripper. We call this the "accumulative error." Using one camera input image, the average accumulative error was 0.070 mm while the gripper was not grasping anything and 0.77 mm while it grabbed objects in the object set. The two image input neural network had an average accumulative error of 0.40 mm while not grasping anything, and 1.43 mm while grasping objects in the object set. We visually compared our single image input neural net results using Rviz (Fig. \ref{fig:actual_v_predicted}).

Live testing results showed that the neural network using a single image input generally outperformed the neural network that used two image inputs. We noted that the last two angles had larger error using the latter neural net. This outcome was unexpected, because we believed that the two image inputs would provide more holistic information than one camera input would. In the camera images we used for the singular input, the end of the soft finger was not visible; however, the mechanical constraints on the finger might allow the neural net we trained to make generalizations of the valid positions the finger joints can have relative to one another. It is still plausible that using two image inputs can give us better results if we change the architecture of the neural network we use. 

Overall, the maximum accumulative error while gripping is small (1.71 mm distance, $5.21^{\circ}$) and is comparable to, if not better than, human hand proprioception. For comparison, a previous study found that while human subjects were localizing their fingertip in 2D space, the group average of the absolute distance between perceived and actual fingertip position was 8.0 cm with a standard error of $\pm{1.0}$ cm \cite{fuentes_wherearm}. The average inference times for proprioception is 7.1 ms (single image input) and 7.3 ms (double image input). The experiment is done on a single TITAN X (PASCAL) GPU.

\myparagraph{Object and Size Classification} We tested each of our eight objects 10 times with the gripper to calculate the accuracy of our size estimation neural network. Out of the 80 test trials, only three objects were incorrectly classified as the wrong size. All incorrect trials occurred while the gripper was grasping the two largest boxes. We believe this could have been caused by a non-firm grip on the objects. 

We also had 100\% accuracy in box/cylinder classification using tactile information.

\section{CONCLUSION AND DISCUSSION}

In this paper, we present a novel exoskeleton-covered soft finger with embedded vision sensors used for simultaneous proprioception and tactile sensing, and we demonstrate that the vision sensors have high-resolution tactile and shape sensing capabilities. The CNN we trained for proprioception of the finger had over 99\% accuracy on our testing set (maximum angle error of the six joint angles was within 1$^\circ$). During live testing, the average accumulative error while gripping is small (0.77 mm distance, 2.22$^\circ$), which is better than human finger proprioception. 

The gripper, which is composed of two of these GelFlex fingers, utilizes proprioception and tactile sensing via multiple neural networks. Out of 80 "bar stock" test trials, only three objects were incorrectly classified as the wrong size, while all were classified as the correct shape (box/cylinder). 

However, some of the current limitations of the robotic gripper include an inability to perform 3D reconstructions using tactile data 
and to grasp a multitude of objects. 
Moreover, we simplified our problem by using only six joint angles, and did not fully capture the complexity of our finger. 

Future work involves improving the robotic gripper so that it can grasp different types of objects and improving the proprioception and tactile sensing algorithms. We would also like to utilize vision-based sensors to estimate more complex finger configurations (i.e., twisting or multi-bending, which classic strain/force sensors cannot easily estimate).


\section{ACKNOWLEDGEMENTS}

This work is financially supported by the Toyota Research Institute. The authors would also like to thank Branden Robert Romero and Shaoxiong Wang for their help in hardware manufacturing and algorithm development, and Achu Wilson for his help in setting up the two Raspberry Pi camera system. 

\addtolength{\textheight}{-10cm}   




\bibliographystyle{IEEEtran}
\bibliography{Ref}
\end{document}